# Robust Unsupervised Small Area Change Detection from SAR Imagery Using Deep Learning

Xinzheng Zhang[a,b,*], Hang Su[a], Ce Zhang[c,d,*], Xiaowei Gu[e], Xiaoheng Tan[a,b], Peter M. Atkinson[c,f,g]

[a] *School of Microelectronics and Communication Engineering, Chongqing University, Chongqing, 400044, China*

[b] *Chongqing Key Laboratory of Space Information Network and Intelligent Information Fusion, Chongqing, 400044, China*

[c] *Lancaster Environment Centre, Lancaster University, Lancaster, LA1 4YQ, United Kingdom*

[d] *UK Centre for Ecology & Hydrology, Library Avenue, Lancaster, LA1 4AP, United Kingdom*

[e] *Department of Computer Science, Aberystwyth University, Aberystwyth, SY23 3DB, United Kingdom*

[f] *Geography and Environmental Science, University of Southampton, Highfield, Southampton SO17 1BJ, UK*

[g] *Institute of Geographic Sciences and Natural Resources Research, Chinese Academy of Sciences, 11A Datun Road, Beijing 100101, China*

## Abstract:

Small area change detection from synthetic aperture radar (SAR) is a highly challenging task. In this paper, a robust unsupervised approach is proposed for small area change detection from multi-temporal SAR images using deep learning. First, a multi-scale superpixel reconstruction method is developed to generate a difference image (DI), which can suppress the speckle noise effectively and enhance edges by exploiting local, spatially homogeneous information. Second, a two-stage centre-constrained fuzzy *c*-means clustering algorithm is proposed to divide the pixels of the DI into changed, unchanged and intermediate classes with a parallel clustering strategy. Image patches belonging to the first two classes are then constructed as pseudo-label training samples, and image patches of the intermediate class are treated as testing samples. Finally, a convolutional wavelet neural network (CWNN) is designed and trained to classify testing samples into changed or unchanged classes, coupled with a deep convolutional generative adversarial network (DCGAN) to increase the number of changed class within the pseudo-label training samples. Numerical experiments on four real SAR datasets demonstrate the validity and robustness of the proposed approach, achieving up to 99.61% accuracy for small area change detection.

# 1. Introduction

Remotely sensed change detection focuses on identifying land-cover changes by analysing the images observed over the same scene at different times (Huang et al., 2011; Li et al., 2020; Cao et al., 2020). In particular, remotely sensed synthetic aperture radar (SAR) imagery is adopted widely for change detection, owing to its ability to penetrate cloud cover, and its insensitivity to atmospheric and lighting conditions (Wang et al., 2016; Gong et al., 2017; Li et al., 2019). Change detection using SAR imagery has been applied in a variety of real-world settings, such as for earthquake damage assessment (Brunner et al., 2010), forest mapping (Pantze et al., 2014) and flood monitoring (Kim et al., 2020). Over the past decades, tremendous effort has been made to develop automatic change detection methods using multi-temporal SAR images. Amongst them, machine learning is currently considered as the most promising and evolving set of approaches (Gong et al., 2017). In general, machine learning-based change detection can be divided into supervised and unsupervised approaches (Li et al., 2019; Geng et al., 2019). The major issue in relation to the supervised approach is the lack of ground reference data, and it often involves manual labelling processes that are labour-intensive and time-consuming (Saha et al., 2020). Thus, unsupervised approaches are employed widely in this field (Li et al., 2015; Jia et al., 2016). The major components of unsupervised approaches include: 1) image pre-processing, 2) difference image (DI) generation, and 3) analysis of the DI and the classification of pixels into changed and unchanged classes (Li et al., 2020; Wang et al., 2020).

The pre-processing step is used to co-register the SAR images (Lei et al., 2019) and filter speckle noise, including multiplicative noise (Gong et al., 2012; Gong et al., 2016; Li et al., 2018). DI generation is a critical step which provides guidance for post-processing. A common approach for DI generation is exploiting spatial information in a local window to reduce speckle noise, such as the mean ratio (MR) and the neighbourhood-based ratio (NR) (Inglada et al., 2007; Gong et al., 2012). However, both MR and NR use a fixed window size for all pixels, and regular windows are adopted to characterise local spatial information. This strategy can smooth the changed area or blur spatial details (such as edges) within the window, where changed pixels are challenging to detect, particularly for small area changes (Wang et al., 2020). To close this gap, we propose a novel method for DI generation based on multi-scale superpixel reconstruction (MSRDI), which exploits the local spatial information within a superpixel (i.e., an image object) instead of a regular window. The mean and median operators are used to reconstruct each superpixel together with a filtered log-ratio DI, enhancing the discrimination between changed and unchanged areas. Multi-scale information is extracted in superpixel segmentation and DI reconstruction followed by fusion to achieve the final MSRDI.

DI analysis is a crucial step for achieving change maps, where clustering algorithms are used to discriminate changed and unchanged pixels. Fuzzy *c*-means (FCM) is a widely adopted clustering approach in change detection using SAR imagery (Gong et al., 2012; Gao et al., 2019; Li et al., 2019). Classical FCM clustered the Gabor feature vectors of the DI in SAR-based change detection (Li et al., 2015). To control the sensitivity to noise, a fuzzy local information *c*-means (FLICM) algorithm was developed by incorporating local spatial information and grey-level characteristics (Krinidis et al., 2010). Gong et al. (2012) reformulated the FLICM (RFLICM) using coefficient of variation to replace Euclidean distance, and employed it for change detection in SAR images. Tian et al. (2018) proposed an edge-weighted FCM by introducing a piecewise smooth prior to balance the trade-off between noise reduction and edge preservation. Li et al. (2019) developed a spatial FCM (SFCM) algorithm with a spatial function added into the fuzzy membership for noise suppression. These FCM-based clustering algorithms are designed mainly to reduce speckle noise while retaining detailed spatial information. However, for small area change detection, these FCM-based algorithms may fail to produce sufficiently accurate results and changed pixels may be misclassified as unchanged pixels. The major problem of current FCM algorithms is the overall loss optimisation objective, resulting in forcing the cluster prototype of the minority (changed) class to migrate to the majority (unchanged) class. This is an inherent problem of these algorithms for imbalanced data clustering. Here, we propose a novel two-stage centre-constrained FCM algorithm (TCCFCM) suitable for grouping imbalanced data. The first-stage of TCCFCM is to identify the preliminary and reliable clustering centres of the changed and unchanged classes. These cluster centres are used as constraints to build the objective function at the second-stage, and to prevent incorrect migration of cluster prototypes.

Recent research suggests that the DI should be divided into three categories: high-probability changed, high-probability unchanged, and an intermediate class (Li et al., 2019; Kalaiselvi et al., 2020). The intermediate class represents pixels that are difficult to discriminate by a specific clustering algorithm. Subsequently, a deep learning classifier with strong discriminative capability can be adopted to identify the intermediate class as either changed or unchanged with high accuracy. A convolutional neural network (CNN), as a patch-based classifier with deep structures, was established to discriminate the changed or unchanged classes in SAR image change detection in Li et al. (2019). Based on the CNN, a convolutional-wavelet neural network (CWNN) was proposed to replace the pooling layer with a wavelet-constrained pooling layer, which can suppress the interference caused by speckle noise effectively (Duan et al., 2017). The CWNN requires a large number of training samples to achieve excellent classification accuracy. However, in small area change detection, pseudo-label training samples of the changed class are often insufficient with limited accuracy. One way to address this issue is through data augmentation (Shorten et al., 2019). For example, a simple linear generation strategy was proposed for data augmentation in SAR image change detection using CWNN (Gao et al., 2019). However, simple linear generation is challenging to increase the generalisation capability of deep learning models (Frid-Adar et al., 2018). Generative adversarial networks (GANs) are powerful deep networks for training sample generation (Goodfellow et al., 2014; Shorten et al., 2019). The GAN model consists of two networks (generator and discriminator) that are trained in an adversarial fashion, where the generator creates fake images and the discriminator discriminates between real and fake images. Radford et al. (2015) further developed a Deep Convolutional GAN (DCGAN), where both generator and discriminator were composed of deep CNNs. The adversarial pair in DCGAN can learn a hierarchy of representations both from image parts and entire images, making it superior to simple linear generation methods in terms of data augmentation (Frid-Adar et al., 2018). Here, we used DCGAN to generate pseudo-label training samples of the changed class with high diversity, while maintaining the main characteristics of the original samples.

In summary, we identified the problem of small area change detection as insufficiently addressed in previous research. This paper, therefore, aims to solve the problem of small area change detection in multi-temporal SAR images through a robust unsupervised small area change detection (RUSACD) approach with deep learning. The major contributions include 1) a novel MSRDI generating method was developed, which can enhance the quality and separability of the DI; 2) TCCFCM was proposed for analysing MSRDI, coupled with a parallel clustering strategy to divide the pixels of MSRDI into three classes: changed, unchanged and intermediate classes; 3) a DCGAN was applied to increase the number of training samples of the changed class and, thus, achieve a balance of training samples amongst classes. This was followed by a CWNN to differentiate the intermediate class into changed and unchanged classes. The proposed approach was tested comprehensively on four real SAR image datasets to validate the model capability. The proposed method has the potential to be applied in a wide range of applications and it can be extended easily given its modular design.

## 2. The proposed approach

The proposed change detection approach RUSACD (Figure 1) includes four parts: 1) pre-processing to de-noise and enhance spatial features; 2) generating the difference image based on multi-scale superpixel reconstruction; 3) parallel TCCFCM clustering to divide the pixels into the changed, unchanged and intermediate classes; 4) applying DCGAN to enrich the pseudo-label training samples of the changed class. A CWNN is trained to allocate the intermediate class to the changed or unchanged classes. For clarity, the pixels that are classified into the changed and unchanged categories by the parallel TCCFCM are expressed as simple pixels because they can be discriminated easily. Intermediate pixels are named hard pixels due to the difficulty in discrimination by clustering only.

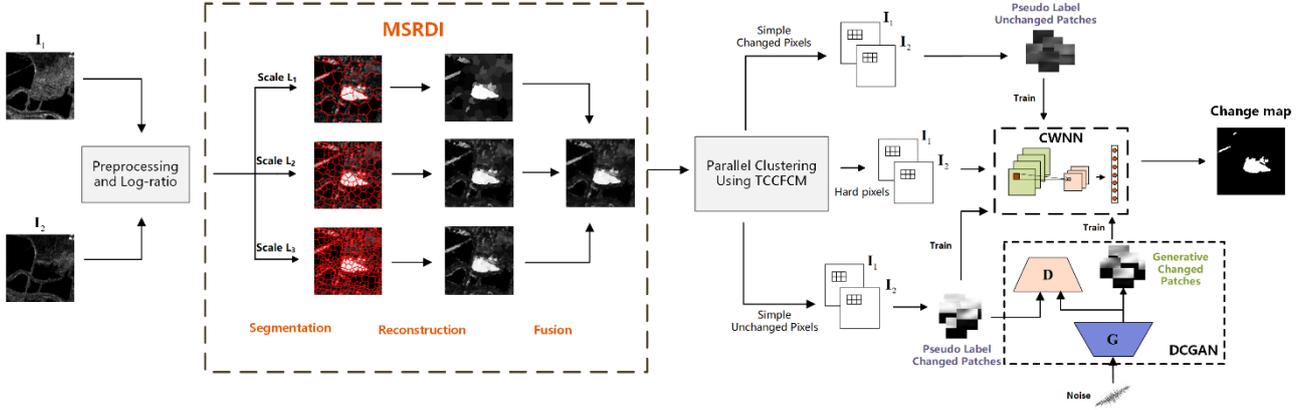

Fig. 1. Flowchart illustrating the proposed RUSACD methodology.

## 2.1 Pre-processing

A weighted average filter is designed for pre-processing, which can suppress noise and enhance spatial domain features effectively. This filter $\mathbf{W}^\eta$ is defined as a matrix, where $\eta$ is an odd number. Each element value of the matrix is determined by the distance between the current position and the centre of the matrix, as shown in Eq. (1). The centre of the matrix is defined as $w_{(\eta+1)/2,(\eta+1)/2} = 2/\eta^2$. The proposed filter maintains edges while suppressing speckle noise compared to other filters such as the mean filter, and retains the continuity of the image.

$$w_{ij} = \frac{1}{\eta^2 \sqrt{(\frac{\eta+1}{2}-i)^2 + (\frac{\eta+1}{2}-j)^2}}; \; i=1,2,...,\eta; j=1,2,...,\eta; \qquad (1)$$

Given multi-temporal SAR images $\mathbf{I}_1$ and $\mathbf{I}_2$, where $\mathbf{I}_1, \mathbf{I}_2 \in \mathbb{R}^{N_r \times N_c}$, the convolution of each SAR image using $\mathbf{W}^\eta$ is taken to acquire two images $\mathbf{I}_1^w(\eta) = \mathbf{I}_1 * \mathbf{W}^\eta$ and $\mathbf{I}_2^w(\eta) = \mathbf{I}_2 * \mathbf{W}^\eta$, where $*$ denotes the 2-D convolution operation.

## 2.2 Generating DI based on multi-scale superpixel reconstruction

Simple Linear Iterative Clustering (SLIC) is used as the basis for superpixel segmentation. A multiscale superpixel reconstruction (MSRDI) is designed here for DI generation. The formal steps for MSRDI include:

Step 1: A log-ratio operator is used to generate the primary DI, and the log-ratio image $\mathbf{I}_{LR}$ is calculated as $\mathbf{I}_{LR} = \left| \log(\mathbf{I}_2^w / \mathbf{I}_1^w) \right|$.

Step 2: The filter template $\mathbf{W}^\eta$ is used to smooth the speckle in the log-ratio (LR) image $\mathbf{I}_{LR}$. The processed log-ratio image is described as $\mathbf{I}_{SLR}(\eta) = \mathbf{I}_{LR} * \mathbf{W}^\eta$.

Step 3: SLIC is used to split $\mathbf{I}_{SLR}$ to obtain $L^t$ superpixel sets that are denoted as $\{S_{l^t}\}_{l^t=1}^{l^t=L^t}$, where $t$ is the index of the scale and $l^t$ and is the index of the superpixel. Each pixel can identify the corresponding superpixel in $\{S_{l^t}\}_{l^t=1}^{l^t=L^t}$. The median is an indicator to evaluate the level of a superpixel value, which is resistant to speckle noise. The mean contains homogeneous information of the corresponding superpixel. Both median sets $\{p_{l^t}\}_{l^t=1}^{l^t=L^t}$ and mean sets $\{o_{l^t}\}_{l^t=1}^{l^t=L^t}$ of $\{S_{l^t}\}_{l^t=1}^{l^t=L^t}$ are obtained, respectively. The superpixel-based DI is reconstructed pixel-by-pixel as

$$I_{SRDI}(i,j) = \alpha_1 I_{i,j} + \alpha_2 p_{l^t} + \alpha_3 o_{l^t} \qquad (2)$$

where $I_{i,j}$ is the pixel value at position $(i,j)$ of $\mathbf{I}_{SLR}$, corresponding to the superpixel $S_{l^t}$.

Step 4: Superpixel segmentation and DI reconstruction are carried out at multiple scales to exploit multi-scale spatial context information, The MSRDI is generated by fusing superpixel-based DIs in $T$ scales as

$$\mathbf{I}_{MSRDI} = \frac{1}{T} \sum_{t=1}^{T} \mathbf{I}_{SRDI}^t \qquad (3)$$

A one-dimensional signal is designed to demonstrate how MSRDI can restrain speckle noise and enhance edges (Figure 2). Compared to other DI generation methods, the MSRDI (Figure 2(d)) has the following advantages: (1) the gap between pixels within a class (changed or unchanged) tends to be small, (2) a significant distinction is present between the two classes, and (3) the boundaries are enhanced and the details are well preserved.

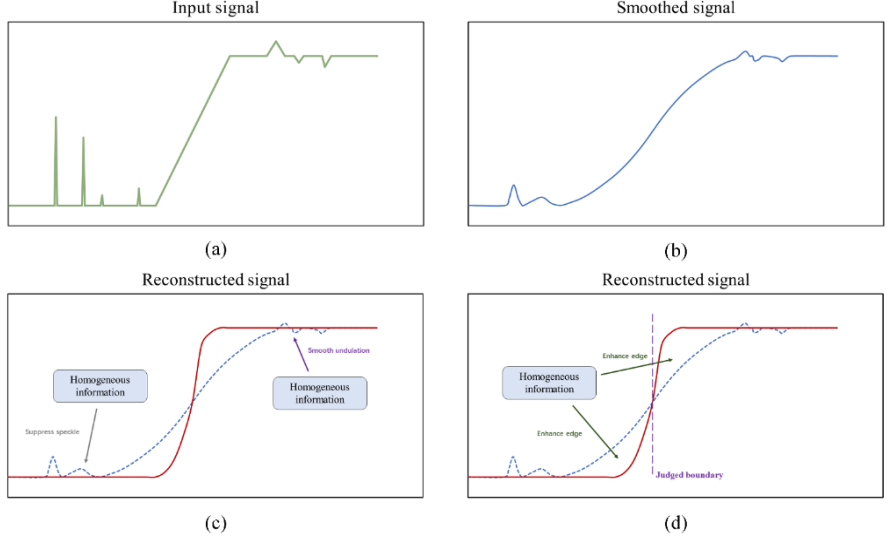

Fig. 2. Diagram using a one-dimensional signal to illustrate how MSRDI can minimise speckle noise and enhance edges. (a) The input signal with speckle and distortion. (b) The smoothed signal using filter $\mathbf{W}^\eta$. (c) The reconstructed signal, in which noise is suppressed and distortion is mitigated. (d) The reconstructed signal, where the edges are enhanced.

## 2.3 Two-stage centre-constrained FCM

A two-stage centre-constrained FCM algorithm (TCCFCM) is designed for unbalanced data clustering to solve the small area change detection problem (Figure 3). The first stage of TCCFCM is to find the preliminary and reliable clustering centres. The most important feature information provided by the DI is the pixel values: high-intensity pixels represent the changed class and low-intensity pixels refer to the unchanged class. Therefore, we select the top $N_p$ samples with prominent high-intensity $\{x_{n_p}^h\}_{n_p=1}^{N_p}$ and top $N_p$ samples with low $\{x_{n_p}^l\}_{n_p=1}^{N_p}$ values. These samples are then clustered by FCM to obtain cluster centres $v_c^{pre}$. Those cluster centres are used as constraints in the second stage to prevent incorrect transfer of cluster prototypes (Fig. 3 (c) and 3(d)). The second stage of TCCFCM performs clustering on all samples $\{x_n\}_{n=1}^{N_r N_c}$. The objective function is defined as:

$$J = \sum_{c=1}^{2} \sum_{n=1}^{N_r N_c} u_{cn}^m \left\| (1-\beta_c)x_n + \beta_c v_c^{pre} - v_c \right\|^2 \tag{4}$$

where $\beta_c$ is a control parameter. We impose a strong constraint by setting a large value for the minority control parameter $\beta_1 (c=1)$, and a small value for the majority control parameter $\beta_2 (c=2)$ to impose a weak constraint. These two parameters are unified as $\beta_1 = \beta$ and $\beta_2 = 0.7 \times \beta$.

The element $u_{cn}$ of the membership partition matrix $\mathbf{U}$ and the cluster centre is derived as:

$$u_{cn} = \frac{1}{\sum_{j=1}^{2}\left(\frac{\left\|(1-\beta_c)x_n+\beta_c v_c^{pre}-v_c\right\|^2}{\left\|(1-\beta_j)x_n+\beta_j v_j^{pre}-v_j\right\|^2}\right)^{1/(m-1)}} \tag{5}$$

$$v_c = (1-\beta_c)\frac{\sum_{n=1}^{N}u_{cn}^m x_n}{\sum_{n=1}^{N}u_{cn}^m} + \beta_c v_c^{pre} \tag{6}$$

where the initial membership partition matrix ($\mathbf{U}^1$) is set randomly at the first stage, and the initial membership partition matrix ($\mathbf{U}^2$) at the second stage is derived by the clustering centre $v_c^{pre}$.

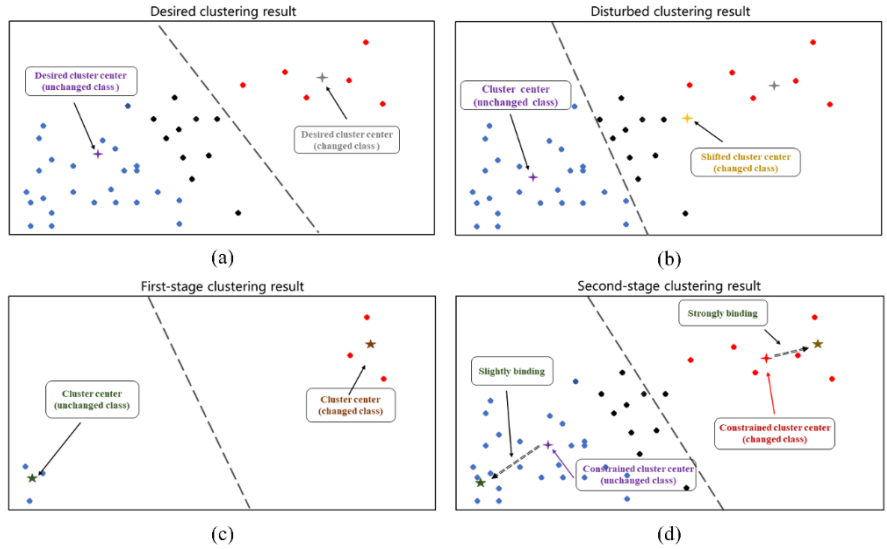

Fig. 3. Illustration of the principle of TCCFCM clustering. The red dots, blue dots and black dots represent pixels or samples of the changed class, low-intensity pixels or samples of the unchanged class, and medium-intensity pixels or samples of the unchanged class respectively. In (c), we obtain reliable clustering centres in the first stage. The reliable centres then constrain cluster prototypes to increase accuracy in the second stage, as shown in (d).

## 2.4 Parallel clustering strategy

A parallel clustering strategy based on TCCFCM is designed to discriminate *simple pixels* including those pixels of both changed and unchanged classes, and *hard pixels* representing pixels of the intermediate class. In the parallel clustering strategy, the sigmoid function is employed to perform two nonlinear mappings on the MSRDI. One prefers the changed class, and the other prefers the unchanged class. A Gabor wavelet transform is used for feature extraction on the two mapped MSRDIs (Li et al., 2015; Gao et al, 2016), implemented by convolving $\mathbf{I}_{MSRDI}$ with a set of Gabor kernels in eight directions on $\gamma$ scales. The maximum response is considered as a feature at this scale. The Gabor feature vector is expressed as $\mathbf{z} = [z_0, z_1, \ldots, z_{\gamma-1}]$. The details of the parallel TCCFCM clustering are as follows:

Step 1: Normalize and centralize the MSRDI $\mathbf{I}_{MSRDI}$.

Step 2: Apply two sigmoid mappings to $\mathbf{I}_{MSRDI}$ with two different parameter sets at each pixel to achieve $\mathbf{I}_1^M = \text{sigmoid}(\mathbf{I}_{MSRDI}; \mu_1)$ and $\mathbf{I}_2^M = \text{sigmoid}(\mathbf{I}_{MSRDI}; \mu_2)$, where $\mu_1, \mu_2$ represent two parameter sets. The sigmoid function is given by Eq. (7).

$$sigmoid(x; \mu) = \frac{1}{1 + e^{-(x+\mu)}} \quad (7)$$

Step 3: Apply Gabor feature extraction to $\mathbf{I}_1^M$ and $\mathbf{I}_2^M$. Gabor feature vector sets are obtained, comprising $\mathbf{Z}^1 = [\mathbf{z}_1^1, \mathbf{z}_2^1, \ldots, \mathbf{z}_{N_r N_c}^1]$ and $\mathbf{Z}^2 = [\mathbf{z}_1^2, \mathbf{z}_2^2, \ldots, \mathbf{z}_{N_r N_c}^2]$, where $\mathbf{z}_i^j$ ($i=1,2,\ldots,N_r N_c; j=1,2$) represents a Gabor feature vector.

Step 4: TCCFCM is utilised to perform two-class clustering on $\mathbf{Z}^1$ and $\mathbf{Z}^2$ to obtain two label sets $\mathbf{Y}^1 = [y_1^1, y_2^1, \ldots, y_i^1, \ldots, y_{N_r N_c}^1]$ and $\mathbf{Y}^2 = [y_1^2, y_2^2, \ldots, y_i^2, \ldots, y_{N_r N_c}^2]$, respectively, where $y_i^j$ represents a label corresponding to a Gabor feature vector. The value of $y_i^j$ is either 0 or 1. The simple average operation is used to encode the two label sets, obtaining the final label set $\mathbf{Y} = [y_1, y_2, \ldots, y_s, \ldots, y_{N_r N_c}]$, where $y_i = (y_i^1 + y_i^2)/2$.

Step 5: If $y_i = 1$, the corresponding pixel is assigned to the changed class $\omega_c$. If $y_i = 0$, the corresponding pixel is assigned to the unchanged class $\omega_{uc}$. The others represent *hard pixels* assigned to the hard class $\omega_h$.

## 2.5 CWNN Deep learning and DCGAN

To increase the accuracy of change detection for a small area, a deep CWNN is constructed to classify *hard pixels*. This is augmented by a DCGAN to increase the sample size for the rare changed class. A patch of size $\lambda \times 2\lambda$ is selected, centred on a pixel of the changed class pseudo-label in the SAR image $\mathbf{I}_1$. Another patch is acquired in the same position from the SAR image $\mathbf{I}_2$ in the same way. Both patches are concatenated into a new patch with $2\lambda \times 2\lambda$ size as a pseudo-label training sample of the changed class. All pseudo-label training samples are produced according to the above method, denoted as $\{\mathbf{P}^c_{\tau_1}\}_{\tau_1=1}^{\tau_1=N_1}$ and $\{\mathbf{P}^{uc}_{\tau_2}\}_{\tau_2=1}^{\tau_2=N_2}$, where $\mathbf{P}^c_{\tau_1}, \mathbf{P}^{uc}_{\tau_2} \in \mathbb{R}^{2\lambda \times 2\lambda}$, representing the changed and unchanged patches, respectively. Hard patches $\{\mathbf{P}^h_q\}_{q=1}^{q=N_h}$ relative to *hard pixels* are obtained in the same way. Thereafter, a DCGAN is used to enrich the pseudo-label training samples of the changed class. Random noise is fed into the generator $G(\cdot)$ to generate false images, and the discriminator $D(\cdot)$ is built to distinguish the fake images from the false and real images. Here, a convolutional generative model $G(\cdot)$ is built, with an objective max-min function as:

$$V(G,D) = L^f + L^r$$
$$= \mathbb{E}\left[\log(1-D(G(\varphi)))\right] + \mathbb{E}\left[\log(D(\mathbf{P}^c_{\tau_1}))\right] \quad (8)$$

$$G^* = \arg\min_G \max_D V(G,D) \quad (9)$$

The structure of DCGAN is shown in Fig. 4. The trained generator $G^*$ is used to expand the data to $\{\mathbf{P}^c_\tau\}_{\tau=1}^{\tau=N_T}$, in which $N_T - N_1$ image patches are generated as fake image patches. $2N_T$ samples are selected to form a training dataset $\{\mathbf{P}^c_\tau, \mathbf{P}^{uc}_\tau\}_{\tau=1}^{\tau=N_T}$, where $\mathbf{P}^c_\tau, \mathbf{P}^{uc}_\tau \in \mathbb{R}^{2\lambda \times 2\lambda}$. The corresponding label $\{L^c_\tau, L^{uc}_\tau\}_{\tau=1}^{\tau=N_T}$ is defined by the parallel TCCFCM clustering result.

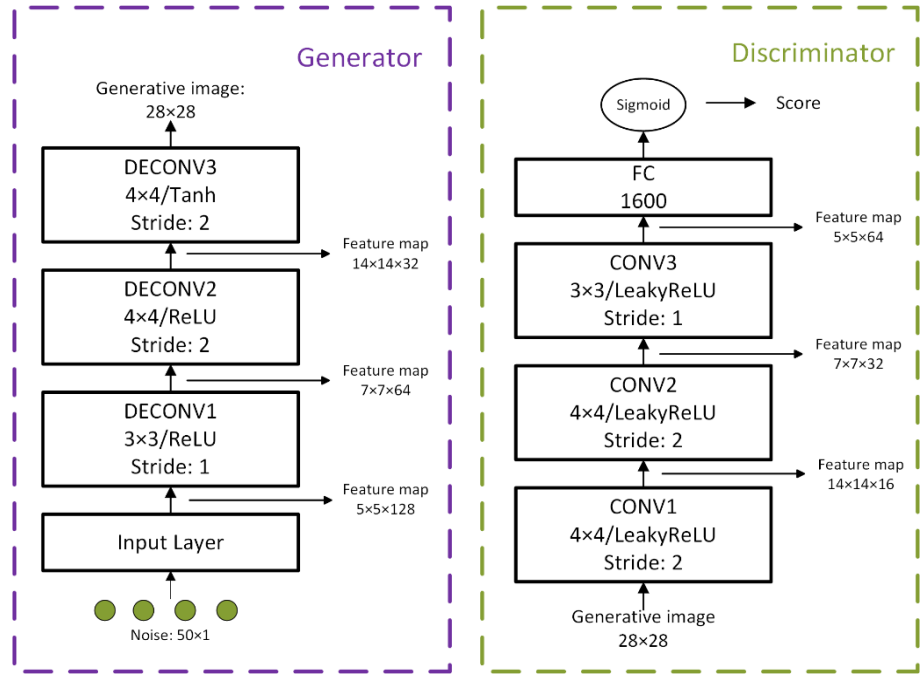

Fig. 4. Network structure of DCGAN. Padding is employed in DECONV2, DECONV3, CONV1 and CONV2.

All the pseudo-label training samples and labels are fed into the CWNN to classify the hard patches. The network structure of the CWNN is shown in Fig. 5, where hard samples $\{\mathbf{P}_q^h\}_{q=1}^{q=N_h}$ are classified into changed and unchanged categories. The final change map can be obtained by combining the labels of *hard pixels* and *simple pixels*.

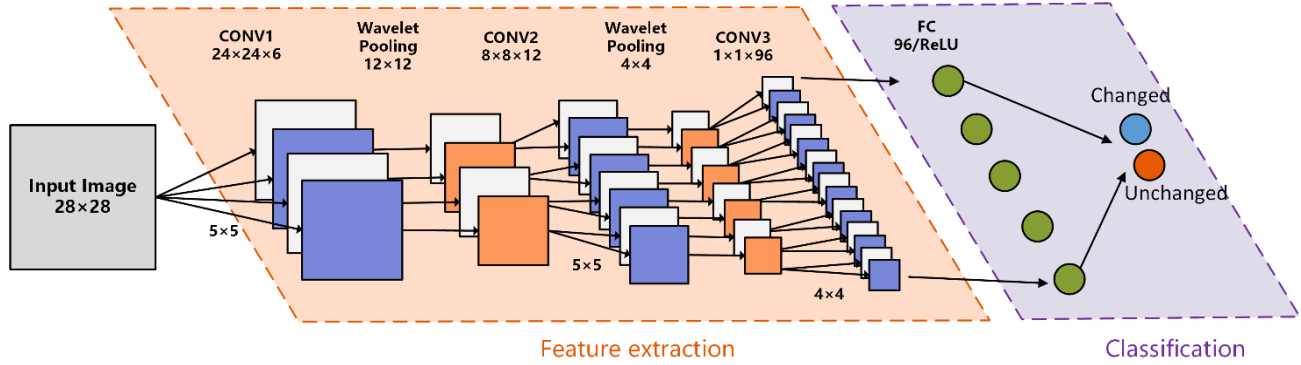

Fig. 5. The network structure of CWNN.

# 3. Numerical experiments

## 3.1 Experimental datasets

Four real multi-temporal SAR datasets (Table 1) were used to evaluate the performance of the proposed approach. Three of these four datasets were acquired over the province of Guizhou in China by the COSMO-SkyMed SAR sensor on June 2016 and April 2017. As shown in Fig. 6, the first image pair (dataset A) with ground reference map consists of mountains and a river, the second image pair (dataset B) with ground reference map includes hills, plains and buildings, and the third image pair (dataset C) with ground reference map is mostly plains. The last image pair (dataset D) was acquired over the city of San Francisco, USA by the ERS-2 SAR sensor in August 2003 and May 2004 (Fig. 7). Speckle noise is shown in datasets A, B and C, which is very challenging for change detection. From the round reference maps in Fig. 6-7, it is clear that the proportion of changed pixels is extremely small compared with unchanged pixels. Dataset D is a benchmark in which there is lower noise and the changes cannot be considered as small area changes. This dataset is used to demonstrate the robustness our proposed approach.

Table 1. Summary of the experimental datasets. $N_c$ and $N_{uc}$ refer to the number of changed pixels and unchanged pixels in the ground reference map, respectively.

| Datasets | Size | $N_c$ | $N_{uc}$ | $N_c:N_{uc}$ |
|---|---|---|---|---|
| A | 400×400 | 1066 | 158934 | 1:149 |
| B | 400×400 | 1492 | 158508 | 1:106 |
| C | 400×400 | 3467 | 156533 | 1:45 |
| D | 256×256 | 4685 | 60851 | 1:13 |

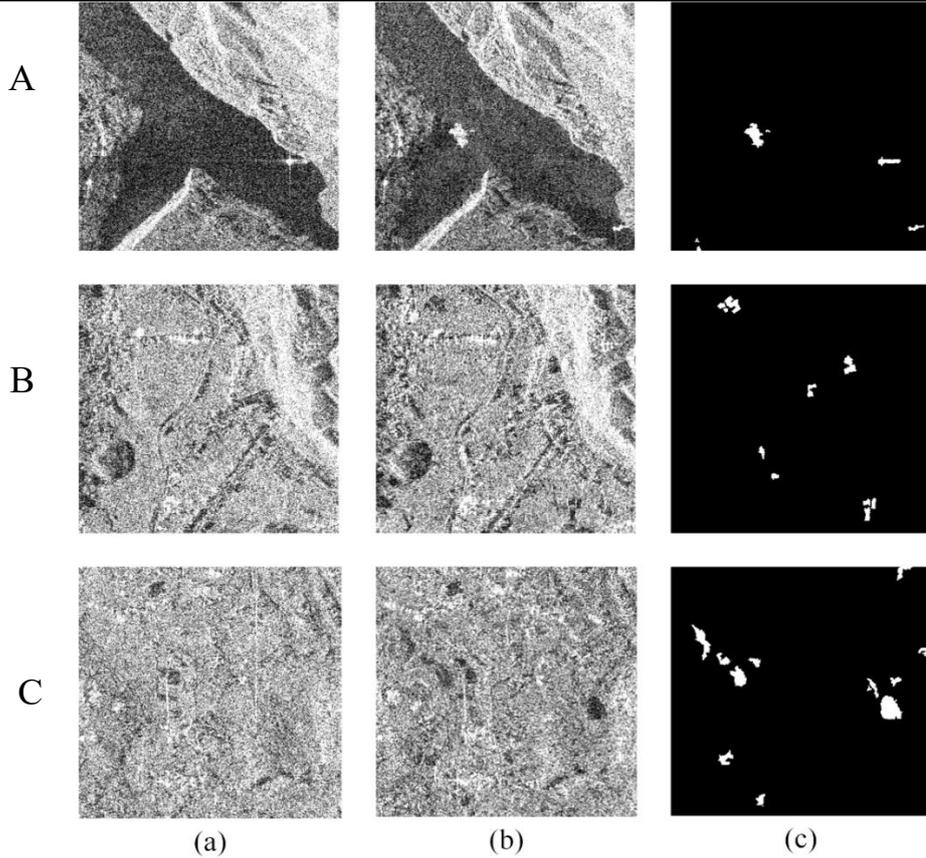

Fig. 6. Dataset A, B and C. (a) Image acquired in June 2016, (b) Image acquired in April 2017. (c) Ground reference map.

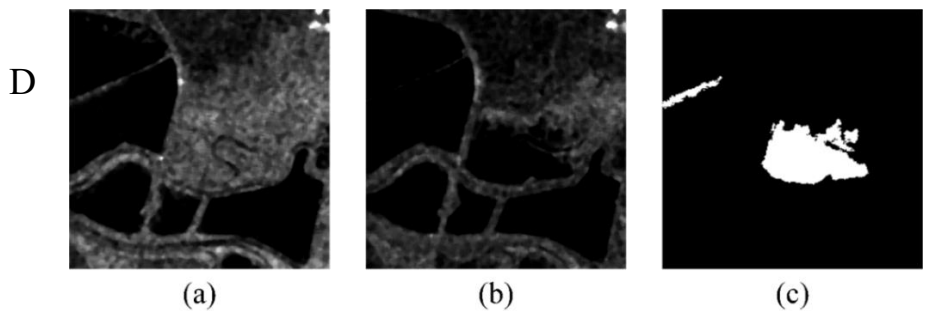

Fig. 7. Dataset D. (a) Image acquired in August 2003. (b) Image acquired in May 2004. (c) Ground reference map.

## 3.2 Experimental setup and parameter setting

The hyper-parameters for each module were set separately. First, the weighted average filter kernel $\mathbf{W}^\eta$ size was set to $\eta = 3$. In MSRDI, superpixel segmentation was conducted on four scales ($m=4$), and $L^1 = 100$, $L^2 = 500$, $L^3 = 1000$, $L^4 = 2000$ were used to obtain spatial information at the different scales. Two sigmoid mapping parameters $\mu_1 = -0.2$ and $\mu_2 = 0.3$ were set for unchanged and changed classes, respectively. $\mu_2$ was larger than 0.2, which ensured the reliable samples fed into the DCGAN. The Gabor feature extraction dimension was $\gamma = 6$, and the constraint parameter of TCCFCM was set as $\beta=0.5$. Finally, the size of the image patch was chosen as $\lambda=14$.

The training of the DCGAN was as follows: 1) the Adam optimiser with a learning rate of 0.0003 was used to optimise the generator, and a learning rate of 0.0006 to optimise the discriminator. 2) Each iteration gave the real image a random value in the range 0.8 to 1 as a score, and the score of the fake image was provided randomly between 0 and 0.2. The number of epochs was 10000, and 640 image patches were selected and divided into 10 batches for training. The training of CWNN used back-propagation with a learning rate of 0.0001.

## 3.3 Evaluation criteria

Five quantitative indicators were used to evaluate the performance of the change detection methods: false alarm (FA), missed detection (MD), percentage correct classification (PCC), Kappa coefficient (KC), and $F_1$ score.

1) **FA**: The number of unchanged pixels that are classified as changed, which corresponds to false positives (FP) in the confusion matrix. The false alarm rate is given by $P_{FA}=FA/(TP+FP)\times 100\%$.
2) **MD**: Number of changed pixels that are not detected, which are false negatives (FN). The missed detection rate is calculated as $P_{MD}=MD/N_c \times 100\%$.
3) **PCC**: Accuracy of pixel-based classification, which can be expressed as
$$PCC = 1-(FA+MD)/(N_c+N_{uc}) \qquad (9)$$
4) **KC**: Kappa coefficient is used for consistency checks, defined as

$$KC = (PCC - PRE) / (1 - PRE)$$

$$PRE = \frac{N_c \times (TP + FN) + N_{uc} \times (TN + FP)}{(N_c + N_{uc})^2} \quad (10)$$

Where TP means true positive, and TN means true negative.

5) **F$_1$**: F$_1$ score is an essential indicator of classification performance, which is defined as

$$F_1 = \frac{2 \times precision \times recall}{precision + recall}$$

$$precision = TP / (TP + FP) = 1 - P_{FA} \quad (11)$$

$$recall = TP / (TP + FN) = 1 - P_{MD}$$

# 4. Results and discussion

## 4.1 Results and comparison

To demonstrate the effectiveness of the proposed RUSACD, four benchmark methods were compared, including: PCA *k*-means (PCAK) (Celik, 2009), neighbourhood based ratio and extreme learning machine (NRELM) (Gao et al., 2016), Gabor feature extraction and FCM with PCANet (GFPCANet) (Gao et al., 2016), and FCM with CWNN (FCWNN) (Gao et al., 2019). We also applied TCCFCM to cluster the MSRDI into two categories (changed and unchanged) as a benchmark MTCCFCM change detection method. The experimental results are shown in Fig. 8, while the quantitative accuracy metrics are given in Table 2. In addition, other methods used for comparison on dataset D (Table 3) include saliency-guided detection with *k*-means clustering (SGK) (Zheng et al., 2017), stacked autoencoder and FCM with CNN (SAEFCNN) (Gong et al., 2017), saliency-guided deep neural network (SGDNN) (Geng et al., 2019) and adaptive generalised likelihood ratio test (AGLRT) (Zhuang et al., 2020).

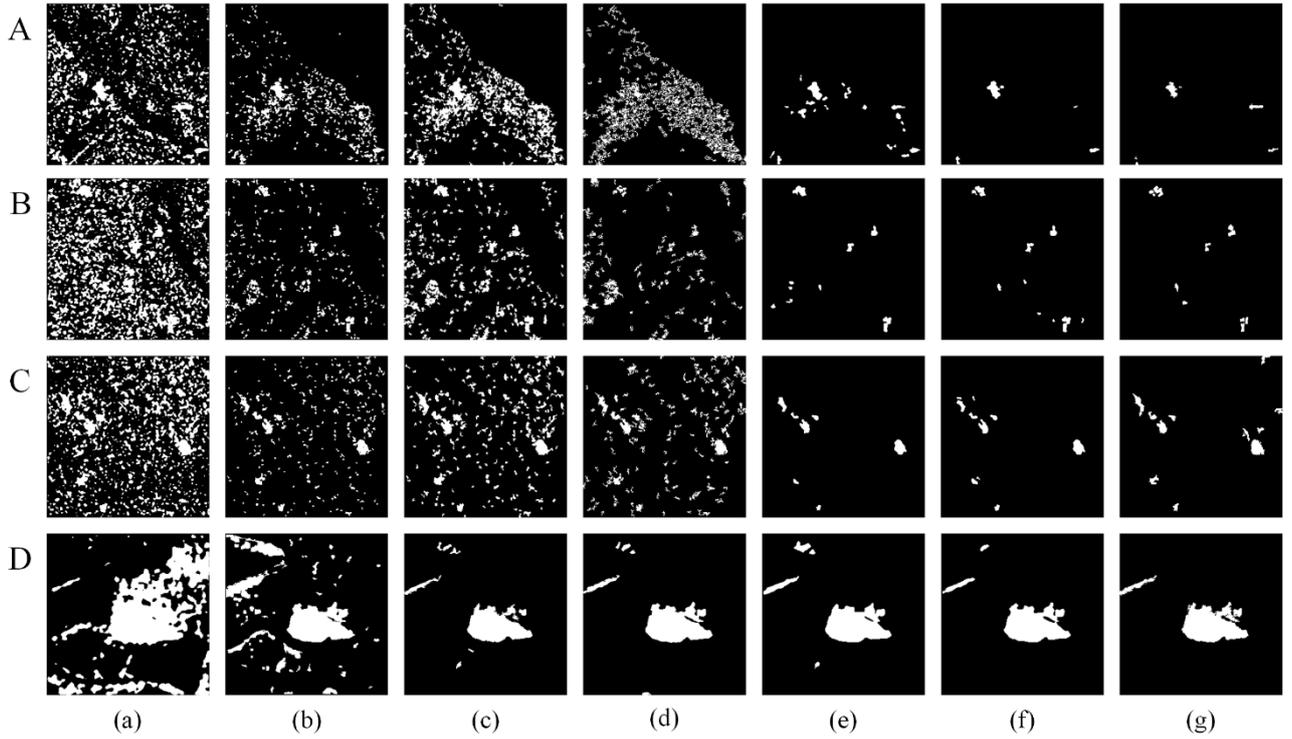

Fig. 8. Comparison of the final change maps on dataset A-D. (a) PCAK. (b) NRELM. (c) GFPCANet. (d) FCWNN. (e) MTCCFCM. (f) RUSACD. (g) Ground reference map.

Table 2. Accuracy metrics for the different change detection methods. Best results are shown in bold.

| | Criterion | FA | MD | $P_{FA}$ (%) | $P_{MD}$ (%) | PCC (%) | KC (%) | $F_1$ (%) |
|---|---|---|---|---|---|---|---|---|
| Dataset A | PCAK | 27822 | **1** | 96.45 | **0.1** | 82.61 | 5.69 | 6.86 |
| | NRELM | 11312 | 42 | 92.00 | 4.10 | 92.90 | 13.74 | 14.76 |
| | GFPCANet | 20935 | 9 | 95.37 | 0.88 | 86.91 | 7.71 | 8.84 |
| | FCWNN | 18108 | 383 | 96.58 | 37.37 | 88.44 | 5.34 | 6.49 |
| | MTCCFCM | 2118 | 107 | 69.76 | 10.44 | 98.61 | 44.68 | 45.21 |
| | **RUSACD** | **390** | 241 | **33.22** | 23.51 | **99.61** | **71.11** | **71.31** |
| Dataset B | PCAK | 43095 | **47** | 96.76 | **3.15** | 73.04 | 4.56 | 6.28 |
| | NRELM | 8066 | 208 | 86.27 | 13.94 | 94.83 | 22.44 | 23.69 |
| | GFPCANet | 14553 | 102 | 91.28 | 6.84 | 90.84 | 14.49 | 15.94 |
| | FCWNN | 7822 | 602 | 73.19 | 17.36 | 94.74 | 38.47 | 40.48 |
| | MTCCFCM | 677 | 330 | 36.81 | 22.12 | 99.37 | 69.45 | 69.77 |
| | **RUSACD** | **596** | 315 | **33.62** | 21.11 | **99.43** | **71.81** | **72.10** |
| Dataset C | PCAK | 31048 | **241** | 90.59 | **6.95** | 80.44 | 13.70 | 17.10 |
| | NRELM | 6527 | 729 | 70.45 | 21.03 | 95.47 | 41.15 | 43.01 |
| | GFPCANet | 11583 | 331 | 78.69 | 9.55 | 92.55 | 32.11 | 34.49 |
| | FCWNN | 7194 | 417 | 87.00 | 27.95 | 95.24 | 20.77 | 22.03 |
| | MTCCFCM | **159** | 1459 | **7.37** | 42.08 | 98.99 | 70.79 | 71.28 |
| | **RUSACD** | 196 | 1406 | 8.68 | 40.55 | **99.00** | **71.53** | **72.01** |

Table 3. Accuracy metrics for the different change detection methods on dataset D. Best results are shown in bold.

| Criterion | FA | MD | $P_{FA}$ (%) | $P_{MD}$ (%) | PCC (%) | KC (%) | $F_1$ (%) |
|---|---|---|---|---|---|---|---|
| AGLRT | **231** | 552 | **5.27** | 11.78 | 98.80 | 91.10 | 91.80 |
| SGK | 561 | 489 | 11.79 | 10.44 | 98.40 | 88.02 | 88.88 |
| SAEFCNN | 231 | 604 | 5.36 | 12.89 | 98.73 | 90.04 | 90.72 |
| SGDNN | 321 | 509 | 7.14 | 10.86 | 98.73 | 90.28 | 90.96 |
| PCAK | 13784 | 182 | 75.38 | 3.88 | 78.69 | 31.40 | 39.20 |
| NRELM | 4466 | **51** | 49.08 | **1.09** | 93.11 | 63.81 | 67.23 |
| GFPCANet | 321 | 337 | 6.88 | 7.19 | 99.00 | 92.42 | 92.97 |
| FCWNN | 545 | 150 | 10.73 | 3.20 | 98.94 | 92.31 | 92.88 |
| MTCCFCM | 621 | 148 | 12.04 | 3.16 | 98.83 | 91.55 | 92.19 |
| **RUSACD** | 279 | 217 | 5.88 | 4.63 | **99.24** | **94.33** | **94.74** |

From Table 2, RUSACD could maintain a change detection accuracy of more than 99%, which is the best among all the methods. The $F_1$ and KC of RUSACD are greater than 70%, which is significantly superior to the other methods. Both PCAK and NRELM produced a low accuracy due to the high sensitivity of *k*-means clustering and extreme learning machine (ELM) to noise. Datasets A, B and C contained massive speckle noise, which led to a very large number of false alarms with limited accuracy. The change detection accuracy for GFPCANet and FCWNN is low since the lack of guiding information in DI and clustering labels.

For dataset D (with low noise), most approaches showed excellent detection capabilities, except for PCAK. For the proposed RUSACD, the PCC reached the highest accuracy of 99.24%, and both $F_1$ and KC exceeded 94%. The MTCCFCM achieved competitive results compared with methods based on deep learning.

## 4.2 Performance analysis of MSRDI

Five DI generation methods were compared in the experiment: LR, SLR (smoothed LR by the proposed weighted average filter), NR, CDI (Zheng et al., 2014) and MSRDI. The results for Datasets C and D are shown in Fig. 9. Compared with the other DIs, the images in MSRDI are smooth, the edges are strengthened and the outliers are suppressed since MSRDI exploits the local spatial information in superpixels at multiple scales. MSRDI has a strong ability to suppress noise and retain clear edges, particularly for small area change detection (the first row of Figure 15).

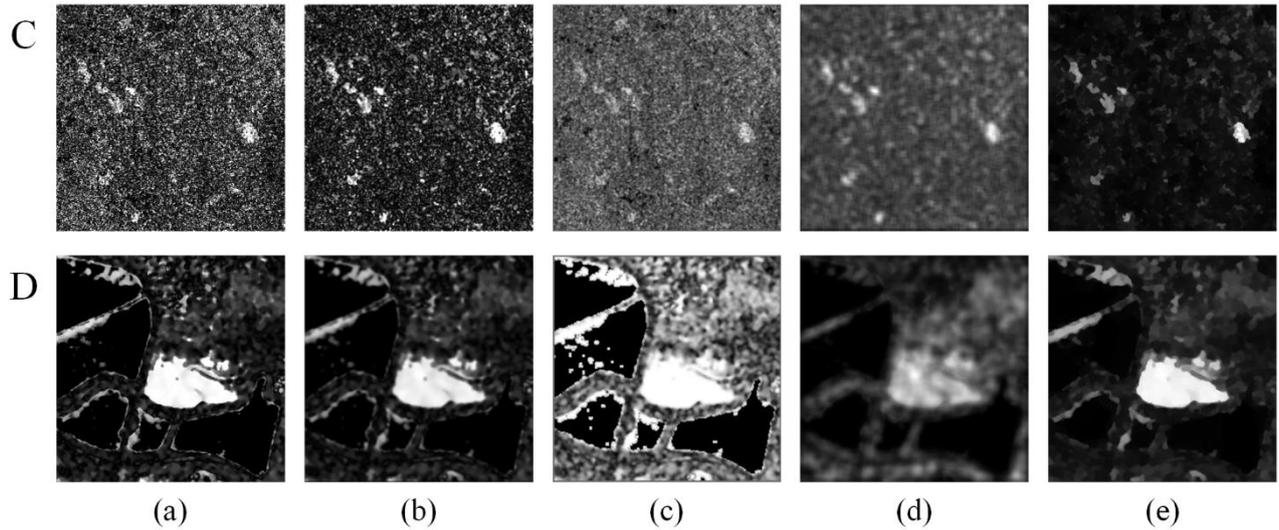

Fig. 9. DIs of the five methods on datasets C and D. (a) LR. (b) SLR. (c) NR. (D) CDI. (e) MSRDI.

The performance of MSRDI was further tested using the Otsu thresholding method to segment all DIs, as shown in Fig. 10 and Table 4. MSRDI is superior to the other methods, with the lowest false alarm rate and the highest accuracy. The NR-based DI and detection results are unsatisfactory for both datasets. For dataset C with strong speckle noise, the mean and median operators in MSRDI were beneficial for suppressing speckle noise. For Dataset D, MSRDI was able to enhance the edges and strengthen the separability of the DI, while smoothing the image.

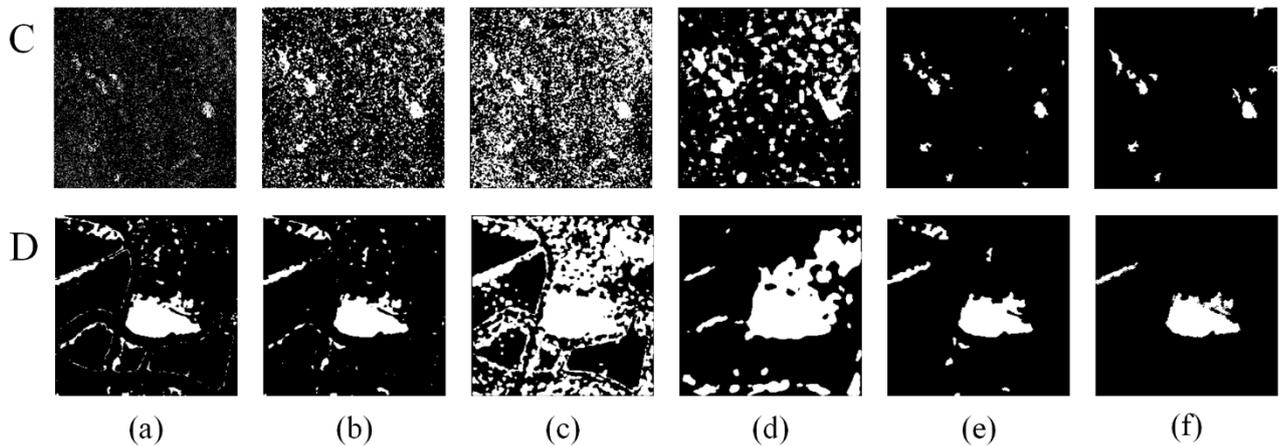

Fig. 10. Change maps of the five DI methods using Otsu thresholding on dataset C and D. (a) LR. (b) SLR. (c) NR. (d) CDI. (e) MSRDI. (f) Ground reference map.

Table 4. Change detection accuracy metrics for five DIs acquired by Otsu thresholding.

| Criterion | Dataset D | | | | | Dataset C | | | | |
|---|---|---|---|---|---|---|---|---|---|---|
| | FA | MD | PCC (%) | KC (%) | $F_1$ (%) | FA | MD | PCC (%) | KC (%) | $F_1$ (%) |
| LR | 2917 | 146 | 95.33 | 72.34 | 74.77 | 12083 | 2000 | 91.20 | 14.28 | 17.24 |
| SLR | 1952 | 130 | 96.82 | 79.71 | 81.40 | 30911 | 309 | 80.49 | 13.42 | 16.83 |
| NR | 23295 | 4 | 64.45 | 18.71 | 28.66 | 54044 | 236 | 66.08 | 6.83 | 10.64 |
| CDI | 11947 | 152 | 81.54 | 35.67 | 42.83 | 24977 | **153** | 84.29 | 17.69 | 20.87 |
| **MSRDI** | **1237** | 174 | **97.85** | **85.32** | **86.48** | **785** | 1110 | **98.82** | **70.72** | **71.33** |

## 4.3 Performance analysis of TCCFCM

Six segmentation methods were used to segment the MSRDI of dataset B to achieve change maps, as shown in Fig. 11 and Table 5. Here, GTCCFCM represents segmentation based on Gabor feature extraction together with TCCFCM. GKM represents the combination of Gabor feature extraction and the *k*-means clustering algorithm. GFCM refers to Gabor feature extraction with FCM clustering. The change map of GTCCFCM produced reliable detection results, with the highest accuracy of 99.37%, demonstrating robustness for small area change detection. Other benchmarks such as FCM produced a significant number of false alarms. The *k*-means algorithm shows some benefits by minimising the incorrect transfer of cluster centres, although it still contains several medium-value unchanged pixels within the cluster prototype of changed class, leading to increased false alarms.

Table 6 shows the accuracy of *simple pixel* classification using the parallel clustering strategy embedded in TCCFCM. The accuracy (PCC) is greater than 99%, showing that the parallel TCCFCM clustering strategy is able to provide credible pseudo-label training samples for deep learning models.

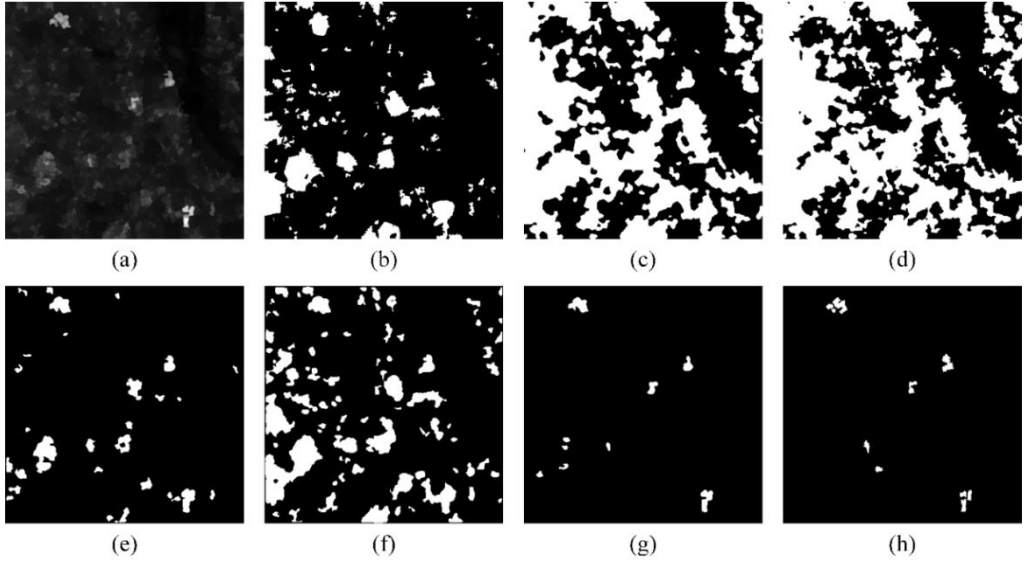

Fig. 11. Change maps of the six segmentation methods on dataset B. (a) MSRDI. (b) Otsu threshold. (c) FLICM. (d) RFLICM. (e) GKM. (f). GFCM. (g) GTCCFCM. (h) Ground reference map.

Table 5. Change detection accuracy metrics acquired by six segmentation methods on dataset B.

| Criterion | FA | MD | $P_{FA}$ (%) | $P_{MD}$ (%) | PCC (%) | KC (%) | $F_1$ (%) |
|---|---|---|---|---|---|---|---|
| Otsu | 17522 | 126 | 92.77 | 8.45 | 88.97 | 11.88 | 13.41 |
| FLICM | 70990 | **3** | 97.95 | **0.2** | 55.63 | 2.24 | 4.03 |
| RFLICM | 78150 | 25 | 98.16 | 1.68 | 51.14 | 1.82 | 3.62 |
| GKM | 5272 | 128 | 79.45 | 8.58 | 96.63 | 32.54 | 33.56 |
| GFCM | 24242 | 32 | 94.32 | 2.14 | 84.83 | 9.14 | 10.74 |
| **GTCCFCM** | **677** | 330 | **36.81** | 22.12 | **99.37** | **69.45** | **69.77** |

Table 6. Simple pixel classification accuracy by the proposed parallel TCCFCM clustering strategy. The $PCC_c$ and $PCC_{uc}$ refer to the accuracy of the changed class and the unchanged class respectively.

|  | $PCC_c$ (%) | $PCC_{uc}$ (%) |
|---|---|---|
| Dataset A | 89.35 | 99.98 |
| Dataset B | 83.71 | 99.82 |
| Dataset C | 99.76 | 99.25 |
| Dataset D | 97.91 | 99.97 |

## 4.4 Performance analysis of deep learning model

Two experiments were conducted on datasets A and D using deep learning (CWNN with DCGAN). In Fig. 12, the red windows indicate that many *hard pixels* are difficult to distinguish in MSRDI, which are classified accurately by the deep learning models. For Dataset D, there are some slight distortions between the bi-temporal SAR images, and they do not represent real land cover changes. RUSACD can distinguish well between the distortions and real changed pixels with the help of deep learning. In dataset A, the proposed RUSACD significantly increased the KC from 44.68% to 71.11%, and $F_1$ from 45.21% to 71.31% compared with MTCCFCM. This is further validated by the three other datasets, showing that CWNN with DCGAN can precisely classify *hard pixels*, thus, increasing change detection accuracy. By contrast, many false alarms are shown when classifying all pixels based on MSRDI and TCCFCM without the deep learning models.

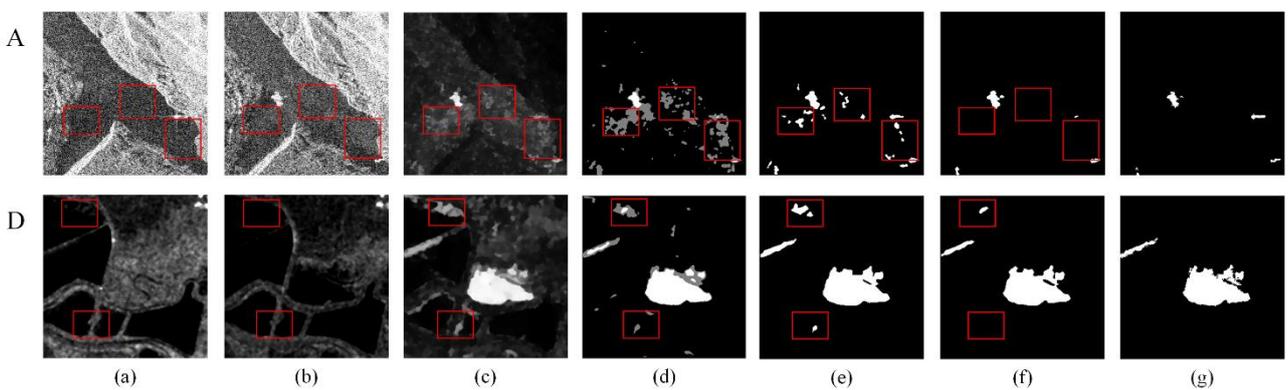

Fig. 12. Analysis of the difficulties in accurately detecting small area changes in dataset A and D. (a) and (b) are the original SAR images. (c) MSRDI. (d) Label map of parallel TCCFCM clustering; white pixels indicate *simple pixels* of the changed class, black pixels represent *simple pixels* of the unchanged class, grey pixels indicate *hard pixels*. (e) Change map by MTCCFCM. (f) Change map by RUSACD. (g) Ground reference map.

The second experiment demonstrated the sample augmentation for small area change detection, where several data augmentation methods were used for comparative analysis, including simple linear generation (SLG) (Gao et al., 2019), ADASYN (He et al., 2008), and our DCGAN method. As shown in Fig. 13, DCGAN was the most effective augmentation method with the highest accuracy in $PCC^H$ and $F_1$ score. The CWNN with sample augmentation can achieve greater accuracy than the benchmark on datasets B, C and D. Although the benchmark experiment results maintain a high accuracy on dataset A, it depends on the distribution of *hard pixels*, which are mostly unchanged pixels in dataset A. The trained CWNN without sample augmentation failed to differentiate *hard pixels*, and classified them all as unchanged pixels.

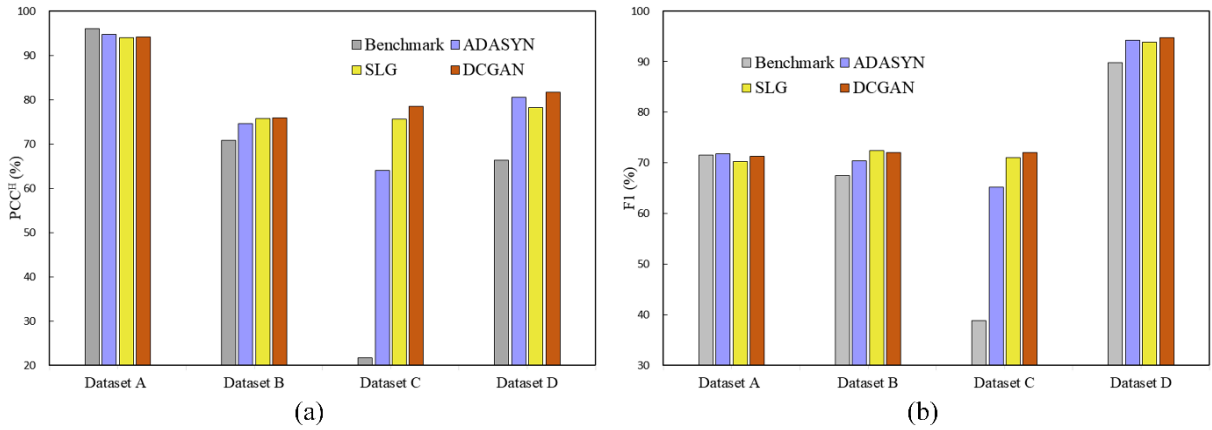

Fig. 13. Comparison of different sample augmentation methods. (a) The classification accuracy of *hard pixels* is the criterion. (b) The *F*1 score of the final change map is defined as the criterion.

## 4.5 Parameter analysis

To test the sensitivity of the results to parameter choice, we investigated the effect of the constraint parameter $\beta$ in TCCFCM on change detection accuracy. We fixed other parameters, and assigned a set of values to $\beta$ as shown in Fig. 14. RUSACD was not sensitive to the parameter $\beta$ for all datasets, and the change detection accuracy was the highest (Fig. 14(b)). The MTCCFCM exhibited stable and satisfactory performance for datasets B, C and D (Fig. 14(a)). The accuracy for dataset A declined when $\beta$ was less than 0.4. These results indicate that RUSACD is robust and accurate compared with MTCCFCM for small area change detection.

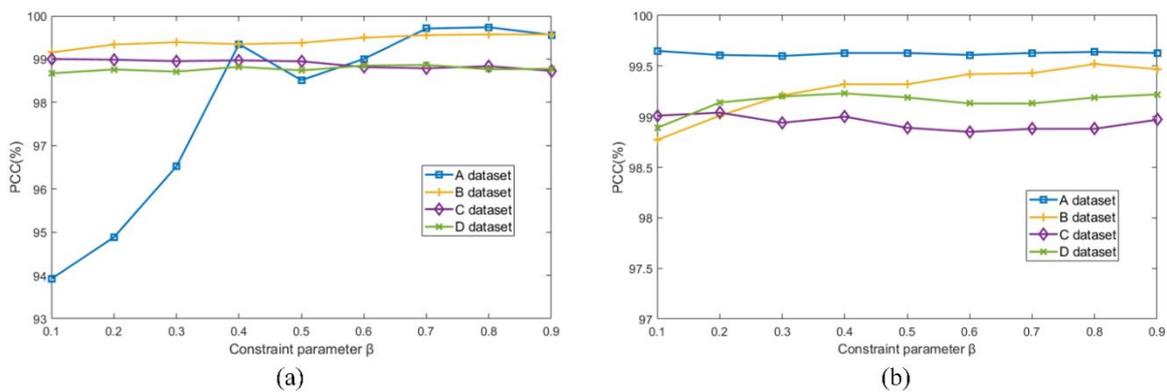

Fig. 14. The relationship between change detection accuracy and parameter $\beta$. (a) Results of MTCCFCM. (b) Results of RUSACD.

## 4.6 Discussion

In multi-temporal SAR image change detection, the speckle noise and image distortion produced in the image acquisition process often cause false changes, leading to a decrease in change detection accuracy. For small area change detection, the number of false changed pixels may be similar to the number of real changed pixels, leading to the problems addressed in this paper. In this research, we proposed RUSACD, as a step-wise modular framework, and demonstrated that it was the most accurate method across a series of state-of-the-art benchmarks for small area change detection.

In the DI generation module, the developed MSRDI aimed to increase the gap between unchanged and changed areas, as well as suppressing speckle noise. It was also able to reduce the interference caused by false-changed pixels.

In the clustering module in the DI analysis, existing FCM-based algorithms usually produce incorrect shifts of cluster prototypes when the number of changed pixels is far less than for unchanged pixels. The TCCFCM was proposed to maintain the appropriate direction of the clustering algorithm through the optimisation process by adopting preliminary centres as constraint terms in the objective function. Furthermore, instead of clustering into two categories, TCCFCM used a parallel clustering strategy, dividing the MSRDI into three classes, which was beneficial for discriminating accurately between false changed pixels and real changed pixels.

In the intermediate pixel classification module, the CWNN was employed for small area change detection. However, as is typical for small area change detection, there were too few training samples of the changed class to train the CWNN. To avoid an imbalance between the changed and unchanged training samples a DCGAN was applied to enrich the training samples of the minority class.

Rigorous evaluation was undertaken of each module in RUSACD by comparing against several benchmark approaches. The experimental results demonstrate the effectiveness and robustness of each module in RUSACD. The proposed method has the benefit that each module can be extended readily and, indeed, the entire framework can be developed as an end-to-end deep network. In future research, the method should be evaluated for a wider range of datasets and applications such as change detection using optical Sentinel-2 images.

# 5. Conclusion

In this paper, a modular RUSACD framework composed of MSRDI, TCCFCM, CWNN and DCGAN was proposed for small area change detection from multi-temporal SAR images. Experiments on four real SAR datasets demonstrated the effectiveness and robustness of each module of RUSACD for small area change detection. The MSRDI in the framework enhances the edges of changed areas and minimises speckle noise and background pixels, increasing the separability between changed and unchanged pixels. The TCCFCM avoids incorrect transfer of clustering prototypes caused by class imbalanced samples, and together with a parallel clustering strategy, the pseudo-label training samples are selected automatically in an unsupervised approach. CWNN obtains discriminative features from the two original SAR images and the MSRDI, while DCGAN further expands the pseudo-label training samples of the changed class, solving the problem of insufficient training samples. We conclude that RUSACD offers many advantages for small area change detection and produced consistently the greatest accuracy of classification amongst the tested benchmarks.

## Acknowledgement

This work was supported by the National Science Foundation of China (61301224) and the Chongqing Basic and Frontier Research Project (cstc2017jcyjA1378).

## Declaration of Competing Interest

The authors declare that they have no conflicts of interest to disclose.